**Implementation and Evaluation of a Gradient Descent-Trained Defensible Blackboard Architecture System**

Jordan Milbrath, Jonathan Rivard, Jeremy Straub
Department of Computer Science
North Dakota State University
1320 Albrecht Blvd., Room 258
Fargo, ND 58108
Phone: +1 (701) 231-8196
Fax: +1 (701) 231-8255
Emails: jordan.milbrath@ndsu.edu, jonathan.m.rivard@ndsu.edu, jeremy.straub@ndsu.edu

**Abstract**

A variety of forms of artificial intelligence systems have been developed. Two well-known techniques are neural networks and rule-fact expert systems. The former can be trained from presented data while the latter is typically developed by human domain experts. A combined implementation that uses gradient descent to train a rule-fact expert system has been previously proposed. A related system type, the Blackboard Architecture, adds an actualization capability to expert systems. This paper proposes and evaluates the incorporation of a defensible-style gradient descent training capability into the Blackboard Architecture. It also introduces the use of activation functions for defensible artificial intelligence systems and implements and evaluates a new best path-based training algorithm.

**1. Introduction**

Multiple forms of artificial intelligence techniques have been developed. Some techniques draw from mathematical optimization principles [1] while others build on inspiration from genetics [2], fictional stories [3], natural phenomena [4] and the behavior of animals [5], [6]. Two well-known techniques are expert systems [7], [8] and neural networks [9].

Rule-fact expert systems are one of the older forms of artificial intelligence, though the term expert system has also been utilized to refer to non-rule-fact systems that embody the knowledge of human experts (e.g., [10]) or provide decision support (e.g., [11]). They trace their origins back to the mid-1960s, when Feigenbaum and Lederberg developed the Dendral system [12]. This system separated the storage of knowledge and its processing [13]. Shortly thereafter, Mycin (which some consider to be the first true expert system) was developed in the 1970s [13]. These classical expert systems utilize a rule-fact network for inference [14], allowing both inductive and deductive reasoning to be implemented.

Another type of artificial intelligence system, neural networks trace their history back to psychologist William James in the late 1800s, according to Eberhart and Dobbins [15]. James discovered several of the key elements of human neurology which underlie artificial neural networks. In the 1940s, McCullock and Pitts further advanced knowledge of how the human brain and neurons, in particular, function and Hebb developed a now well-known method for updating network weightings [15]. In the late 1950s, Rosenblatt implemented the first electronic neural network [15], using an IBM 704 computer, which introduced the perceptron concept [16]. In the 1960s, Widrow and Hoff introduced a computerized neural network [15] featuring a training mechanism [17] based on least mean squares error optimization.

In the intervening decades, both expert systems and neural networks have found numerous uses. Expert systems have been demonstrated for applications ranging from aerospace [18] to education [19]. Neural networks have been used in numerous areas ranging from welding [20] to rainfall estimation [21]. A variety of extensions of both techniques have been proposed such as support for fuzzy logic [22], for expert systems, and speed enhancement [23].

Another type of system, the Blackboard Architecture [24], is based on the rule-fact expert system concept. However, it extends the capabilities of the system from simply providing recommendations to actually making and implementing decisions, thereby altering the operational environment of the system. A simple Blackboard Architecture implementation adds the concept of actions [25] to the rules and facts of an expert system. Notably, numerous variants of Blackboard Architecture implementations have been previously proposed.

Due to the growing prevalence of artificial intelligence systems, concerns have been raised that many techniques – and neural networks, in particular – lack understandability and accountability. Concerns about systems causing discrimination [26], perpetuating historical power structures [27] and learning bias [28] have been raised. The concept of eXplainable Artificial Intelligence (XAI) has been proposed in response to these concerns. XAI systems are either designed to be human understandable or they feature an interpretation mechanism to aid systems' human understandability [29]. Building on this, a "defensible" system was developed which utilized a rule-fact network; this was then optimized using backpropagation [30], [31]. Because the optimization engine couldn't change the network structure, it was limited to learning how to best utilize logically sound data relationships, preventing it from learning spurious correlations, which may be forms of bias or due to non-causal data correlations (which may not apply in all circumstances).

This defensible system was based on an expert system rule-fact network and, thus, lacked actualization capabilities. It also did not include the activation functions used by neural networks. This paper proposes and evaluates a system that expands the gradient descent-trained defensible expert system to be a defensible Blackboard Architecture. It also adds an activation function capability and proposes a best path-based optimization algorithm, building upon the prior work presented in [32].

This paper continues, in Section 2, by presenting prior work which provides a foundation for the work presented herein. In Section 3, the proposed system is presented. Then, Section 4 discusses the experimental design used for this work. Next, Section 5 presents and analyzes data characterizing the efficacy of the system. Finally, the paper concludes, in Section 6, and discusses potential areas of future work.

## 2. Background

The system proposed herein is an optimizing rule-fact network expert system with actualization capabilities, based on the Blackboard Architecture. Machine learning techniques are used for training the network.

This section presents prior work in several areas that provide a foundation for this current work. In Section 2.1, previous machine learning techniques are described, with a focus on those that provided inspiration for the proposed system. Section 2.2 describes expert systems and discusses their benefits and drawbacks. Section 2.3 reviews prior work on gradient descent and its influence on the network in which it is employed. Neural networks are described in Section 2.4. A discussion of the limitations that many current neural network systems have is also included.

## 2.1. Machine Learning

Machine learning and artificial intelligence allow systems to adapt their functionality using the data that is supplied to them.  This opens the door for potential uses that were previously unseen [33]. Machine learning has found applications across a wide range of fields. Due to the wide range of applications for machine learning a variety of machine learning architectures have been built for specific purposes [34]. Small data sets often require simpler machine learning techniques to make sense of them, while large, complex networks of data frequently use more sophisticated forms of machine learning to optimize the processing of their larger quantity of data [35].

There are several different techniques that are used for machine learning. One of these is the use of decision trees. When a decision tree receives an input, the tree looks for the next branch that is most related to the target item. This keeps repeating until a leaf node on the tree is reached, at which point the leaf is returned as a conclusion [36]. Artificial neural networks, which are discussed later, are networks inspired by biological neural networks. They are collections of nodes with weights connecting them. These weights determine the strength of the connection between any two given nodes, which can then be used in evaluation of different queries [37].

Machine learning's utility has been demonstrated within several fields, including the visualization and interpretation of images [38], mortality prediction and prevention for individuals with critical diseases [39], and software development [40]. It has been demonstrated to have particular utility for tasks related to pattern recognition [41] and data analysis [42].

## 2.2. Expert Systems

An application is considered to be within the expert programs class when it states what it knows and the reasons why it knows those facts to be true [43]. This is often based on creator-predetermined system of functionality, commonly in the form of formulas and precise conclusion-finding algorithms. While many programs exhibit qualities of an expert system, not all of them include full transparency as a deliberate aspect of the program.

Expert systems trace their roots back to two key systems: Dendral [12], which was produced in 1965, and Mycin [13], which was produced in the 1970s.  Classical expert systems employed a rule-fact network [14] and a separate inference engine.  Their facts were Boolean true or false values and rules were used to identify other facts that could be asserted as being true, if their inputs were satisfied.

Expert systems have been used for many applications including aerospace [18], engineering [44], land management [45], [46], and agriculture [47], education [19] and robotics [48], [49], in addition to a variety of medical applications [50].

There have been several proposed optimizations for expert systems [51]. One example used algorithms based on evolutionary genetics.  A fitness function was optimized iteratively. For each iteration, a new "generation" of data was created, which is designed to satisfy the fitness function better than the last. From there, a recombination phase was completed, selecting the "fittest" individuals from the data pool [52]. This optimization and many others were developed in an attempt to find the most efficient and effective way to solve a given problem. Expert systems have also been extended to support fuzzy logic [22].

2.3. Neural Networks and their opaque nature

Artificial neural networks stand in stark contrast to classical expert systems. They mirror the biological function of human brain neural networks by combining multiple layers of nodes. Each node in a layer is connected to each node in the adjacent layers with a specific associated weight. Each of these nodes acts like a neuron and, by changing the weights through a training process, the artificial neural network can be trained to recognize complex patterns and solve problems [53].

Artificial neural networks are often used for pattern recognition or classification. They have been used in disease diagnosis, speech recognition, data recovery, and object [53], and facial recognition [54] as well as in industries such as banking [55], construction management [56] and for medical diagnosis [57]. A key drawback to artificial neural networks is their opaque approach. While operators can see the weights attached to each neuron path, they do not know what each neuron represents. This can lead to misclassification and other invalid results, particularly in edge cases that the network was not trained for.

If an artificial neural network does make a mistake, it is hard to identify this or to find the logical cause of a known error, due to the opaque nature of the system [61]. Despite these limitations, neural networks are widely used and have been extended to enhance their operating speed [23] and to resolve issues in areas such as bias factors [58], [59] and initial condition sensitivity [60]. Work has also been conducted to provide attack resilience [61], [62].

2.4. Gradient Descent and Backpropagation

Gradient descent is a popular algorithm used to optimize neural networks. Although there are several different methods for implementing gradient descent within a system, they share the same underlying concepts and produce similar results [63]. In general, gradient descent is used to minimize the output of a function, which is often the error for the goal that the system is trying to achieve [64].

Optimization is performed by influencing objects within the system in a way that moves the result towards a local minimum. A negative slope for the function that is being minimized is followed until the local minimum is reached, at which point a goal has been reached. The approach that is taken for gradient descent implementation often depends on the number of variables that the system developer seeks to influence with each pass. Several factors, such as the group of variables that are influenced and the calculation used to determine the change for those variables, have an impact on the efficacy of gradient descent for any given system [65].

Backpropagation, a form of gradient descent, is used in many neural networks to train the set of nodes. The process starts at the output node and works backwards, basing each previous node's weight value on the next node in the network's value. Backpropagation has been used in applications across the realm of machine learning, ranging from detecting tumors [66] to predicting market trends for products [67].

Prior to neural networks, the use of backpropagation was limited, due to the existence of alternative linear methods of descent that guaranteed that global minimums would be reached. However, as complex and multilayered networks became more common, local minima were found to be useful for many applications [68].

Several issues have been identified with backpropagation, such as the speed of learning and the lack of influence of new inputs on a preexisting network [69]. Ultimately, though, several variations of backpropagation have been successful, using the same fundamental concepts.

*2.5. Blackboard Architecture*

The Blackboard Architecture was introduced by Hayes-Roth [24] in 1985; however, it is based off of an earlier system called Hearsay-II [70]. While it is conceptually similar to rule-fact expert systems, it goes beyond their typical capabilities by adding an actuation capability. Basically, this involves adding a third node type (the action) to the typical rule-fact network [25].

Despite its conceptual simplicity, this additional capability facilitates the system's use for numerous applications that require the system to be able to affect the operating environment. The Blackboard Architecture has been shown to be useful for tutoring [71], robotic [72], [73] and vehicle command [74] and modeling proteins [75], among other applications. To support these and other uses, a variety of enhancements have been proposed to the base Blackboard Architecture concept.

Enhanced capabilities include Goodman and Reddy and, separately, Craig's proposed use of pruning to enhance system speed [76], [77]. Velthuijsen, Lippolt and Vonk proposed a parallel [78] Blackboard Architecture concept. Ettabaa, et al. suggested the use of distributed Blackboard Architecture [79] processing. The use of boundary nodes [80] for distributed processing was also proposed, building upon this concept. Craig proposed techniques for message handling and filtering [81]. Techniques for approaching the Blackboard Architecture from a solution-oriented (Blackboard solving) perspective [82], [83], for use in goal-driven Blackboard Architecture implementations, have also been proposed.

**3. System Overview**

To characterize the efficacy of gradient descent in a Blackboard network, an evaluation system was developed. This section presents an overview of the proposed system, which builds upon previously implemented Blackboard systems, adding a gradient descent training capability and using a scalar (non-Boolean) representation of facts. While previous Blackboard networks simply used rules to connect facts, the proposed system also assigns weights to each fact in each rule, allowing for more complex training and contribution calculation. This section begins with a logical overview of the system. Then, system operations are described. Finally, the process of training the system is presented.

*3.1. Basic Structure*

System operations start with the creation of a rule-fact network. For this system, a fact is a node that stores a value between 0 and 1 (inclusive). This value can be representative of a proposition's probability of being true in a real-world scenario. Alternately, it can represent a degree of applicability of a concept or a level of group membership. Rules connect these facts. Each rule is connected to two input facts and one output fact. If the rule is executed, it sets the output fact based on an equation and the input facts' values. Complex rule structures can be made using chains of rules.

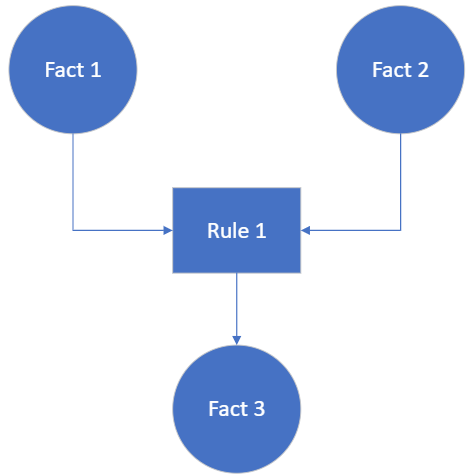

**Figure 1.** Basic Rule-Fact Network Structure.

The basic rule-fact structure can be repeated as many times as a system developer wants. A single fact can be an input for multiple rules, and each fact can also be the output of multiple rules.

For each of its input facts, a rule has an associated weight, which is a value between 0 and 1 that is used in the calculation of the output of that rule. The weightings of the two input facts must sum to 1. Facts that serve as the input for several rules will have distinct weights for each of those rules. When running the network, an output is calculated using the input fact values and weights for that rule. A rule's output is calculated by taking the sum of the products of each input fact with its corresponding weight. This is shown in Equation 1, where $F_1$ is the value of input fact 1, and $F_2$ is the value of input fact 2. $W_1$ is the weight associated with fact 1 for the rule, and $W_2$ is the weight associated with fact 2. $R_{output}$ is the calculated output for the rule. The equation is, thus:

$$R_{output} = F_1 \times W_1 + F_2 \times W_2 \qquad (1)$$

In addition to the weights, lower and upper threshold values can also be defined, providing an activation function for the rule. Both of these threshold values must be between 0 and 1 (inclusive), and the lower threshold must be less than the upper threshold. When a rule output is calculated, the result is compared to these lower and upper threshold values. If the result is between the values, the value of the output fact for that rule is set to the result of the calculation. If not, no change is made. Using trigger thresholds provides a method regulating the range of possible output values for a rule, and with that, the possible values that the output fact is set to.

The purpose of this is to allow for more control over the network's functionality. By setting a threshold range, an operator can ensure that an output fact's value stays within a certain range. In addition to this, it may also be useful for an operator to require a result output to be above a certain level to impact the output fact. If a rule's calculation result is too low, it may not be significant enough to warrant changing the value of the output fact. In either case, threshold values can be used to implement the applicable functionality.

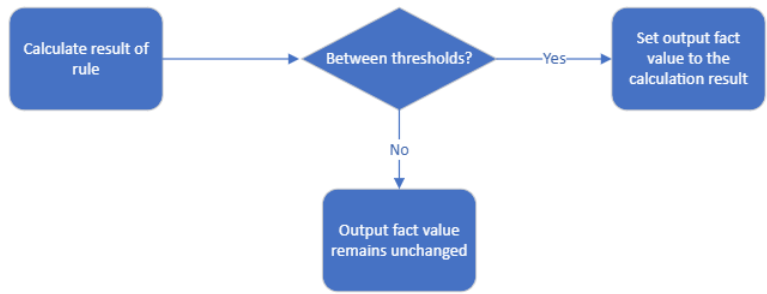

**Figure 2.** Diagram of Rule Threshold Logic.

In addition to the facts and rules, actions can be defined in the network. The addition of actions transforms the system from simply providing decision support to being an actuating system. Actions are commands that are run whenever the result of a particular associated rule's calculation is between the rule's lower and upper thresholds. Since this is also the requirement for setting an output fact value, this means that whenever an output fact value is set by a rule, any actions associated with the rule will be run as well.

*3.2 Running the Network*

Running the system entails the sequential execution of rules between an identified start and end fact. A run begins at a user-identified start fact. The fact's value is set and all rules that have this fact as an input are run in a random order, so that no rule runs first every time. As each of the rules are run, they are added to a list of used rules, which indicates that they have already been run during this iteration and prevents loops which could cause the system to run indefinitely. The output facts for each of the rules are set and are used to find the next set of rules that will be run. As with the first iteration, all of the rules that have any of the output facts as inputs are added to a list. Again, rules are selected and run in a random order until the list is empty. This process repeats until the output facts no longer serve as inputs for any rules that have not yet been run. Once this process is completed, the value of the end fact is returned.

*3.3 Training the Network*

A key feature of the system is the capability to train the rule-fact network. This process optimizes the weightings of the input facts for the ruleset to increase the accuracy of the output of the network, using user-provided training data. To do this, training data is provided to the system, with each entry containing a desired input and a desired output. The network then uses the concept of gradient descent to adjust the weights that are associated with the relevant facts in the rules, thus causing the output of the network with the provided input to produce a result that is closer to the target output.

To train the network, a start fact and an end fact are identified. For the training to have an influence on the network, the start fact must be either a direct or indirect input for a rule that has the end fact as an output. The two facts can have any number of facts and rules between them, and they can be connected via any number of rule paths. Figures 3 and 4 are examples of valid start and end fact selections for training. Figure 5 shows an invalid training pathway where there is no path from the start fact to the end fact.

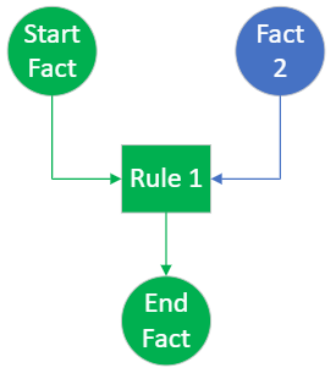

**Figure 3.** Valid Start and End Fact Selection (Example 1).

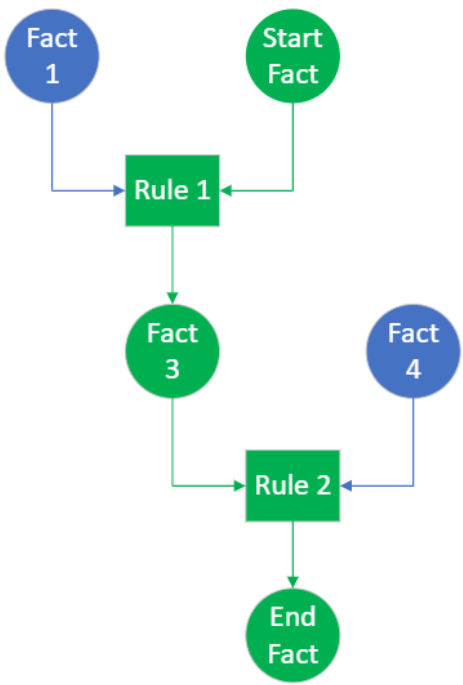

**Figure 4.** Valid Start and End Fact Selection (Example 2).

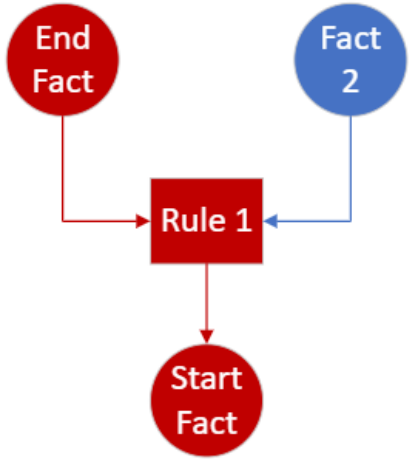

**Figure 5.** Invalid Start and End Fact Selection.

In addition to identifying a start and end fact, a value for the start fact and a target value for the end fact must be provided. While the network is training, it is working to produce the target value for the end node by altering the weightings of the input facts for some or all the rules between the start fact and the end fact. The rule input weightings that are changed are those that comprise the path from the start node to the end node that results in the start node having the highest contribution to the end node. In networks with a high rule-to-fact ratio, there may be many paths from the start fact to the end fact. To find the path that results in the start fact having the greatest contribution, a modified version of Dijkstra's algorithm is used.

*3.3.1. Fact Contribution*

For a given rule, an input fact's contribution to the rule's output fact is proportional to the weight associated with that fact for that rule. Contributions are unidirectional; an input fact contributes to an output fact, but the output fact does not contribute to an input fact via a rule. To calculate the contribution of a start fact that is not directly connected to the end fact via a single rule, the weight of the start fact is multiplied by the contributions of each fact between the start and end facts. The contribution of any start fact, $C_{Start}$, to a path-connected end fact can be determined using the following equation (modified from [31]):

$$C_{Start} = W_{Start} \times \prod W_{IF} \qquad (2)$$

where $W_{start}$ is the weight of the start fact within the first rule between it and the end fact and $W_{IF}$ represents the weight of each intermediate fact, with respect to the rule, of which it is an input fact for along the path from the start to the end fact.

Figure 6 presents an example of this. In this example, fact 6 is the end fact. All of the other facts are potential start facts because they all contribute, either directly or indirectly, to fact 6. Two of the direct contributions to fact 6 are fact 5, with a contribution of 0.4, and fact 1, with a contribution of 0.5. Fact 3 contributes to fact 6 indirectly. To find the contribution of fact 3, all paths between fact 3 and fact 6 are evaluated. In this example, there is only one path, via fact 4. As shown in Equation 2, the contribution of fact 3 is equal to the weight of fact 3 multiplied by the weights of all the intermediate facts. Since the only intermediate fact is fact 4, the contribution of fact 3 will be 0.2 × 0.6, which equals 0.12.

A more complex calculation is required to determine the contribution of fact 2 to fact 6. There are two separate paths that connect fact 2 to fact 6, so the one that results in the greatest contribution will be used. The contribution of fact 2, using the path going via rule 1, is equal to the weight of fact 2 in rule 1, which is 0.5. Using the path going via fact 4, the potential contribution of fact 2 is based on the weightings of fact 2 in rule 2 and fact 4 in rule 3, which is 0.8 × 0.6, and equals 0.48. Since 0.5 is greater than 0.48, the contribution of fact 2 to fact 6 is 0.5. The path going through rule 1 is the most influential path.

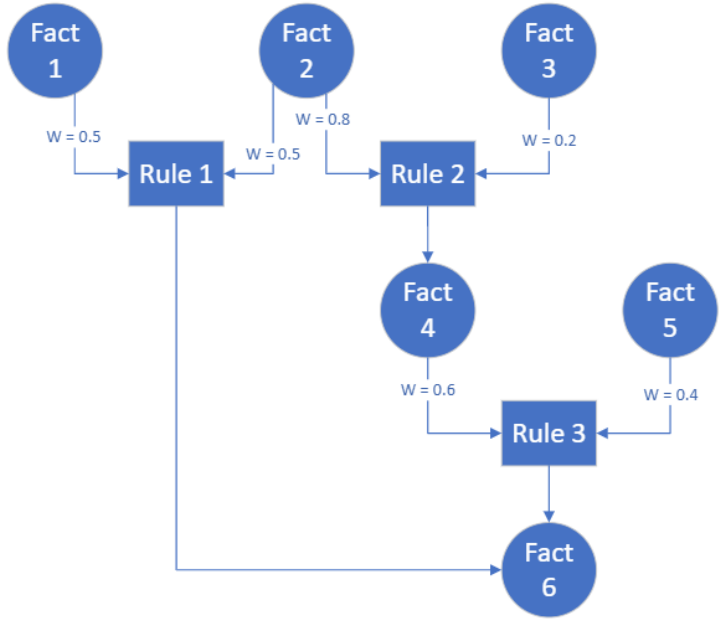

**Figure 6.** Example of a Rule-Fact Network.

*3.3.2. Determining the New Rule Weights*

To train the network, the weights associated with the facts along the most influential path are adjusted. The change that is made to the weights of a particular rule is calculated using the equation:

$$D_i = \frac{C_i}{C_{Total}} \times V \times \Delta_R \qquad (3)$$

In this equation, $D_i$ is the individual difference that will be added to the rule weighting of fact in question. $C_i$ is the contribution of the fact for which the difference is being calculated. $C_{Total}$ is the sum of the contributions of all of the facts along the most influential path from the start fact to, but not including, the end fact. V is a velocity value, which is a positive number between 0 and 1 that determines how much of an impact each training has on the weighting of the rules, thus determining the speed at which the training process changes rule weightings. Finally, ΔR is calculated using the following equation:

$$\Delta_R = \frac{|R_{Current} - R_{Target}|}{MAX(R_{Current}, R_{Target})} \qquad (4)$$

In this equation, $R_{Current}$ is the result of running the network from the start fact to the end fact in its current state. $R_{Target}$ is the user supplied target result that the network is training to achieve. The MAX() function returns the largest of the values that is passed to it.

## 4. Experimental Methodology

To test the efficacy of the system and characterize its performance, data was collected through a process of experimentation. Simulations using the system were run using randomly generated networks, as described in [84]. Seventy combinations of input settings were simulated. The following inputs were configured for each simulation:

| | |
|---|---|
| Fact Count | Velocity |
| Rule Count | Iteration Count |
| Action Count | Rule Weights |
| Trigger Range | Action Fire Probability |

For each experimental condition, 1,000 runs were conducted. However, not all runs completed successfully, due to random network creation or start/end path selections.

The standard parameters that were used for the simulations are shown in Table 1. When testing how a specific input value affects the system's performance, that input value was changed from the standard while leaving the others the same. These changes are described for each experimental condition in Section 5.

Table 1. Standard experimental parameters.

| Input Attribute | Value |
|---|---|
| Fact Count | 100 |
| Rule Count | 100 |
| Rule Weights | 0.5 |
| Action Count* | 0 |
| Action Fire Probability* | 0 |
| Training Iterations | 10 |
| Training Velocity | 0.1 |
| Trigger Index Range | Random with at least 0.5 difference |

*Although tested in several experimental conditions, actions were not used by default in the simulations to reduce random error. When actions are used, the action count standard is 100 and the fire probability standard is 1.

Each test run consists of multiple steps. First, using the given fact, rule, and action counts for the experimental condition, a random rule-fact network is generated. Facts are randomly associated with rules as inputs and outputs and actions are randomly attached to the rules. For experimental purposes, actions introduce uncertainty. When an action is fired, a random fact in the network is selected, and that fact's value is randomized. This models applications where actions introduce new and unpredictable data into the network, which is a worse-case scenario for optimization. Of course, for many applications, the result of actions may not be data production or the data collected may be more foreseeable.

From the generated network, two facts are selected that are connected by rules, meaning that one of the facts contributes to the value of the other fact. A random value is generated and assigned to the start fact, and the network is run. The resultant value of the end fact is used as a target value for training. The network is then trained to move the end fact's value towards the target value, as described in Section 3.3. Between each training iteration, the values of the facts are reset to their values in the initial run.

Training is performed to evaluate the effectiveness of the training process for the given experimental condition. After each training, another run of the network is completed. The result of this run is also recorded. The results from – and the analysis of – these simulations are presented in the subsequent section.

## 5. Data and Analysis

This section considers the performance of the proposed system from a number of perspectives. It characterizes the system's overall performance and demonstrates the comparative efficacy and impact of the changes to the system presented in [31], [32] which are proposed herein.

This section continues with analysis of the impact of changes to the network size in Section 5.1. Then, in Section 5.2, the impact of different levels and velocities of training is analyzed. After that, in Section 5.3, the impact of trigger thresholds is evaluated. Next, the correlation between system runtime and error is assessed, in Section 5.4. Finally, in Section 5.5, the impact of concurrently altering multiple experimental parameters is evaluated.

*5.1. Network Size Analysis*

The first area of analysis is the characterization of the impact of different network sizes on the performance of the system. Table 2 and Figure 7 show how the runtime of the system changes with an increasing number of rules. It was shown, as expected, that as the number of rules increased in the network, so did the average run time for that simulation. This increase was shown to be at a greater-than-linear rate of growth. This result is consistent with the operations of the system, which could (in the worst case) require that every rule in the network be executed. For each test that was performed, an initial run of the network was performed, then training was performed, and a run was performed post-training.

It was shown that both the pre- and post-training runs had roughly the same average run time. This shows that the training process does not notably change the runtime required by the system.

**Table 2.** Average pre- and post-training runtime by rule count level.

| Rule Count | Average Train Time | Average Initial Run Time |
|---:|---:|---:|
| 50 | 376.3246753 | 433 |
| 100 | 3772.323 | 3782.854 |
| 250 | 43084.897 | 42471.48069 |
| 500 | 194079.8892 | 195615.6921 |

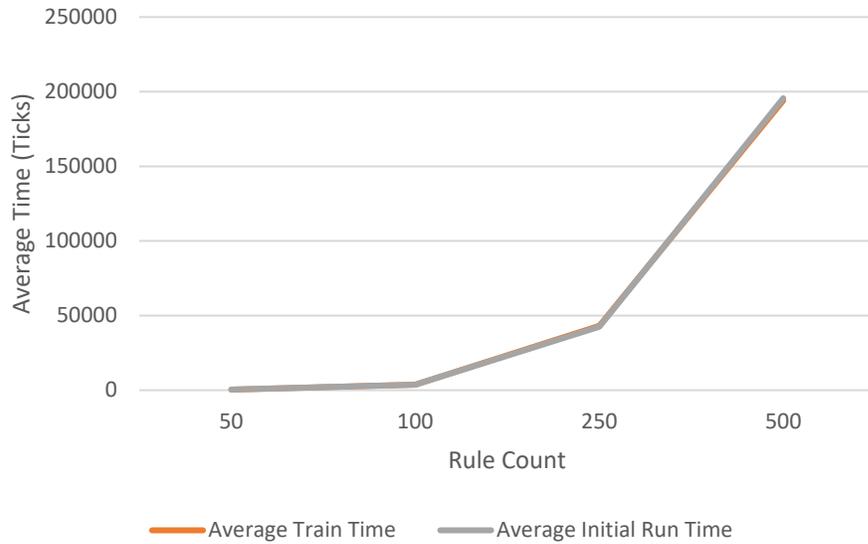

**Figure 7.** Average run time by rule count.

Next, the error rate of the system was assessed with varying numbers of facts. This data is shown in Table 3 and Figure 8. It was found that as more facts were added to the networks, the average error first increased and then slightly decreased. Since there are only three facts connected to each rule – two input facts and one output fact - there is a limit to the number of facts that can be connected in a network given a fixed number of rules. However, with only a fixed number of facts, there can be any number of rules connecting those facts since they can be used in multiple rules, both as inputs and outputs, creating a complex network.

It is important to note that the average error doesn't increase proportionate to the number of facts. This demonstrates the potential suitability of the system for use with complex networks with many facts, allowing the system to be used in networks of greater scale.

**Table 3.** Average error by fact count.

| Fact Count | Average Error |
|---|---|
| 50 | 0.092020588 |
| 100 | 0.128124334 |
| 250 | 0.159895705 |
| 500 | 0.137549715 |

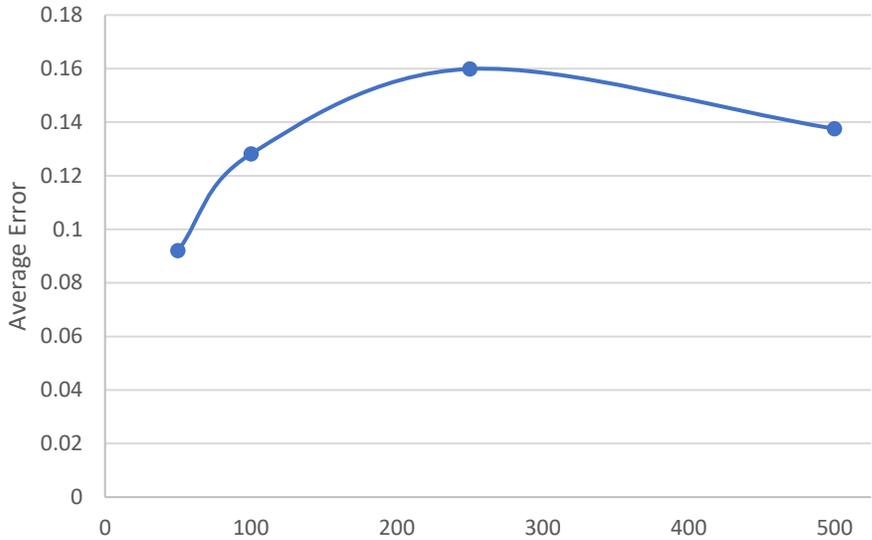

**Figure 8.** Average error by fact count.

As shown in Table 4 and Figure 9, increasing the rule count decreases the error of the network. Notably, a decline in error is seen between the 50 and 500 rule levels; however, the rate of error reduction declines with greater rule counts. This suggests that there may be a point of diminishing return or an optimal range of rule-fact ratios; however, this may differ significantly by application area and the applicable rule configuration for each application. Notably, a greater number of rules allows the network to embody more nuance of a phenomena, potentially reducing the error level.

When there are more rules in a network, there is also a greater potential for multiple paths from the start fact to the end fact and a higher number of rules along those paths. This allows more of a change as data passes through the network. This means that as networks become more complex, so does the training for those networks. Training a single path will have less of an impact on a network with multiple connections than one with very few; however, more rules provide more locations for optimization and network 'memory', potentially reducing error through greater path specificity.

**Table 4.** Average error by rule count.

| Rule Count | Average Error |
|---|---|
| 50 | 0.185505253 |
| 100 | 0.128124334 |
| 250 | 0.078693022 |
| 500 | 0.0750262 |

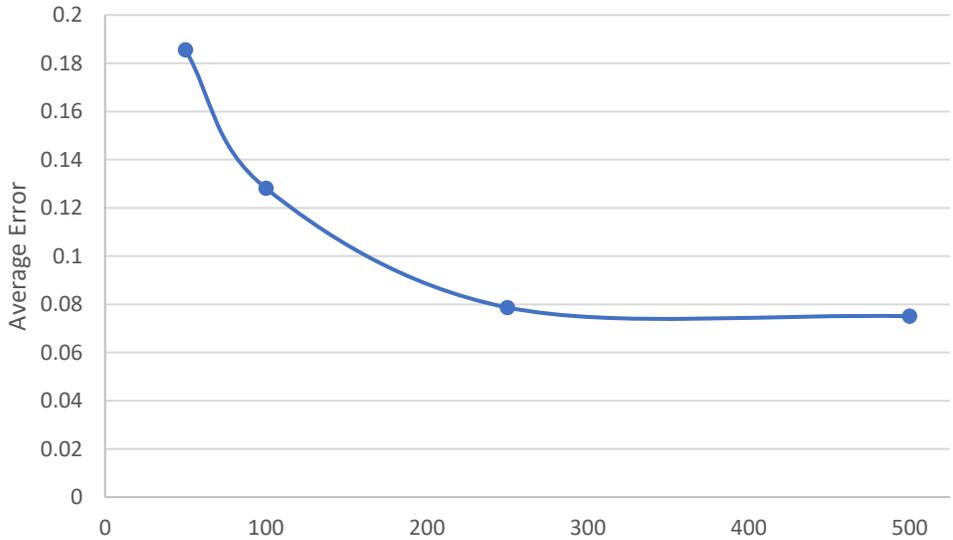

**Figure 9.** Average error by rule count.

Next, focus turns to the impact of the number of action nodes in the network. This data is shown in Table 5 and Figure 10. An increasing number of actions present in a network, when actions inject uncertainty, was shown to result in increased error.

The approach taken for this analysis presumes that the impact of actions running is not known (or readily predictable) a priori. Thus, the actions that were implemented in the simulation change the value of a random fact to a randomly selected value. This is consistent with actions bringing in new, previously unknown information into system operations.

In many cases, especially with a large number of facts in a network, the introduced change is unlikely to be to a fact along a path between the selected start and end facts. Additionally, the changed fact may not be impactful if its value is overridden by the result of a rule for which that fact is an output. A greater number of actions, of course, increases the likelihood of changes' impactfulness.

Notably, the average error level, shown in Table 5, generally increases with the introduction of more actions. However, it shows a slight decrease, at the 500-action level.

**Table 5.** Average error with different levels of actions in the system.

| Action Count | Average Error |
|---|---|
| 0 | 0.128124334 |
| 50 | 0.141485327 |
| 100 | 0.166171066 |
| 250 | 0.182415093 |
| 500 | 0.177010817 |

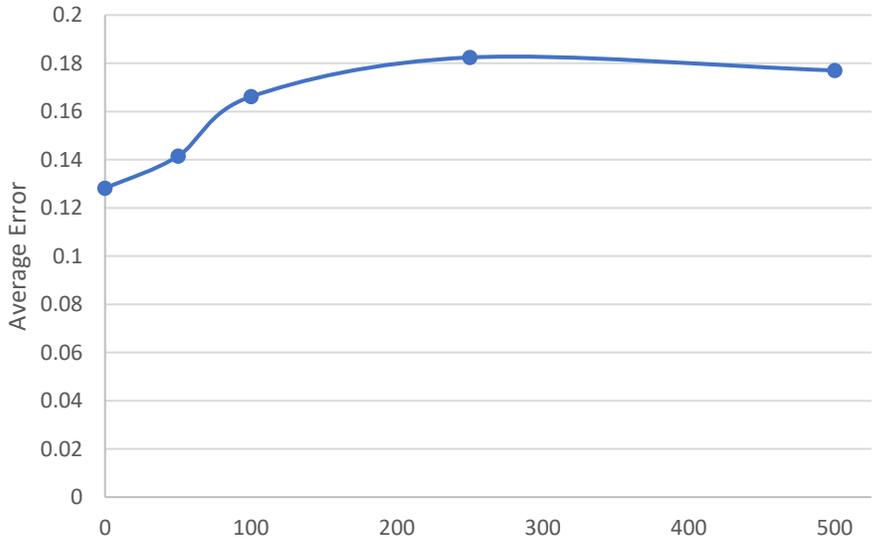

**Figure 10.** Average error by action count.

*5.2. Training Impact*

Focus now turns to assessing the impact of training on system performance. Prior work on gradient descent training of expert systems (as well as the use of similar training for neural networks) suggests that training should have a pronounced impact on system performance.

Simulations were run using 10, 100, 250, and 500 training iterations. These simulations, for which data is shown in Table 6 and Figure 11, do not appear to show a clear trend.  The worst performance occurs at the middle training levels and the lowest training level performs better than the 100 and 250 training iterations levels and almost as well as the 500-iteration level.  This seems to indicate that training actually has a negative impact on performance.

As neural networks have been shown to suffer from over-training [85] and the pattern shown between the 100, 250 and 500 does not have an immediately clear cause, further analysis is needed.

**Table 6.** Average error for different numbers of training iterations.

| Training Iterations | Average Error |
|---|---|
| 10 | 0.128124334 |
| 100 | 0.140812774 |
| 250 | 0.142707331 |
| 500 | 0.127506103 |

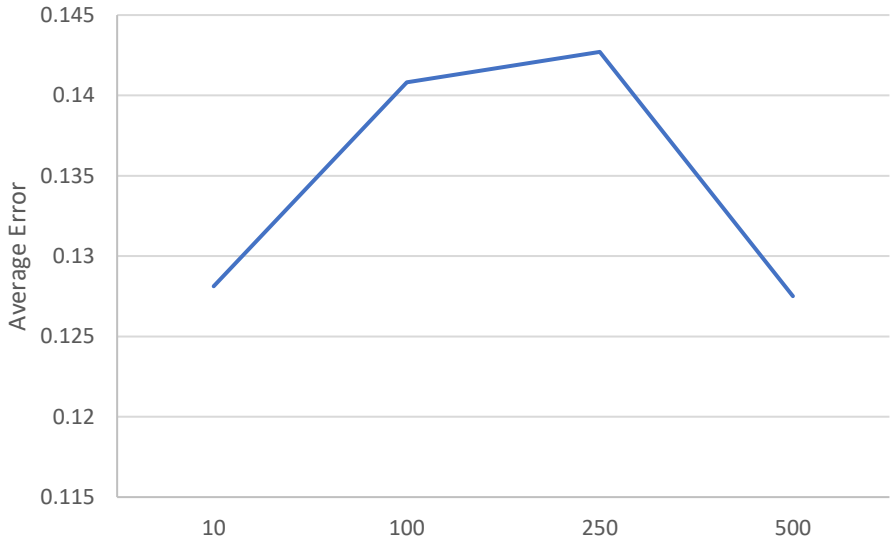

**Figure 11.** Average valid error by training iterations.

To further investigate this, the number of tests that performed best for each group of 25 training iterations within the 100, 250 and 500 iteration training tests was determined. Figures 12 to 14 show the data for each training level. Notably, the 100 iterations data shows peaks at the lowest and highest training levels. However, a dramatically different pattern is shown for the 250 and 500-iteration data. The 500-iteration data pattern explains the other two. It shows that the ideal level of training is near 300 iterations. While a small peaks is present at the highest training level, it is nowhere near as high as the one near 300 iterations. Thus, in the 100 and 250-iteration data, it appears that the peaks at the lowest and highest levels include networks that would outperform their performance at these training levels, if more training was conducted.

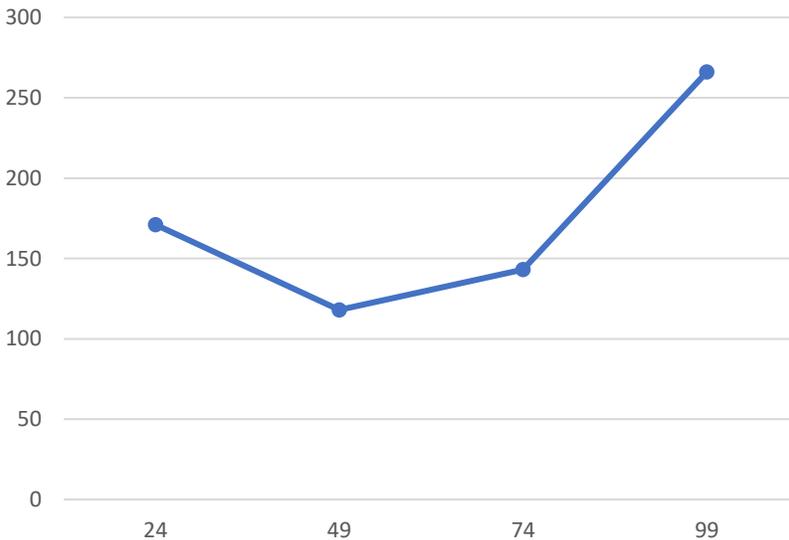

**Figure 12.** Number of best results by training level for 100 training iterations.

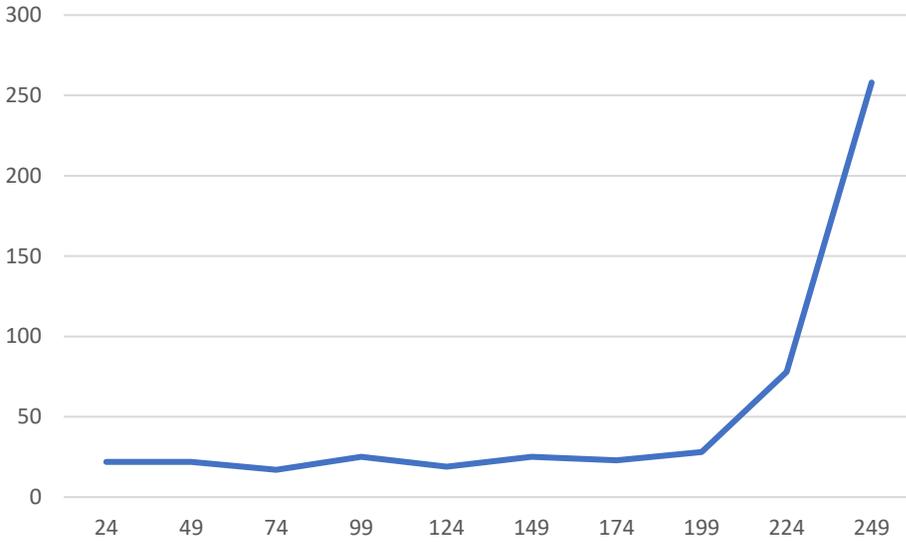

**Figure 13.** Number of best results by training level for 250 training iterations.

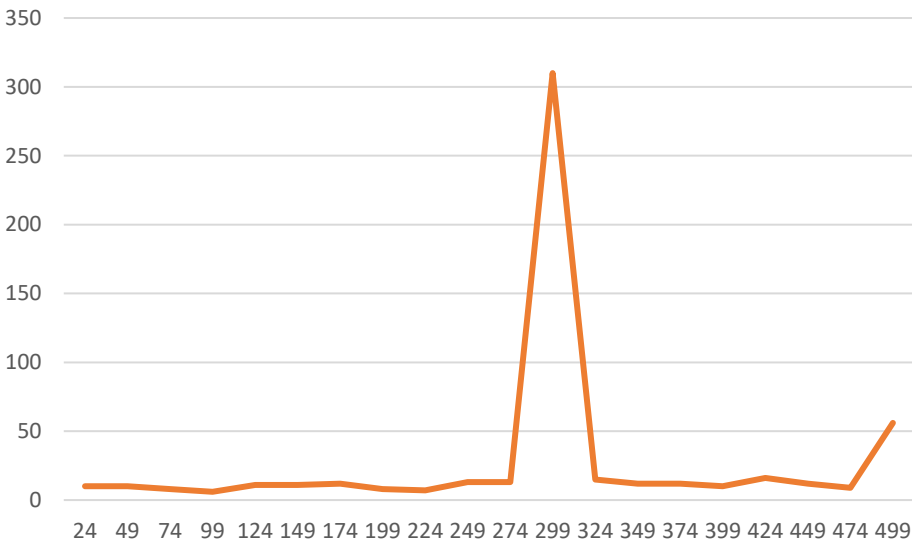

**Figure 14.** Number of best results by training level for 500 training iterations.

The data also demonstrates that the algorithm is susceptible to overtraining and can move away from a better solution to an inferior one. Thus, a mechanism that selects the best result from the entire set of training results can improve its performance. Table 7 shows the impact of adding this capability at each training level. Notably, this significantly improves the results over simply using the result at the end of a given number of training iterations. It is also notable that, when doing this, there is a clear decrease in error between successively higher levels of training.

**Table 7.** Mean and median best result by training level.

| Training | Mean | Median |
|---|---|---|
| 100 | 0.1171 | 0.0816 |
| 250 | 0.0303 | 0.0100 |
| 500 | 0.0113 | 0.0007 |

As the system can move from a local minima through a higher error level to a better (potentially also local) minima, simply stopping if performance declines does not maximize performance. Future work could identify ways to improve the process of reviewing results to select the best performing, ideally using a test during each training iteration.

Focus now turns to training velocity. Training velocity can also have a notable effect on system performance, as it causes each training iteration to be more or less impactful on the system, depending on the velocity value. Through the simulations performed, it was shown that training velocity like the number of training iterations, has a negative correlation with average error (i.e., increased training velocity levels improve system performance).

As shown in Table 8 and Figure 15, every training velocity produced a lower average error than that of the initial 0.025 velocity value. The lowest error level was obtained at the highest training velocity that was tested, 0.25. Error declined with increased velocity between each set of levels. This indicates that increasing the velocity of training increases training effectiveness.

**Table 8.** Average error by training velocity.

| Training Velocity | Average Error |
|---|---|
| 0.025 | 0.149748806 |
| 0.05 | 0.143028564 |
| 0.1 | 0.140812774 |
| 0.2 | 0.134484665 |
| 0.25 | 0.128187673 |

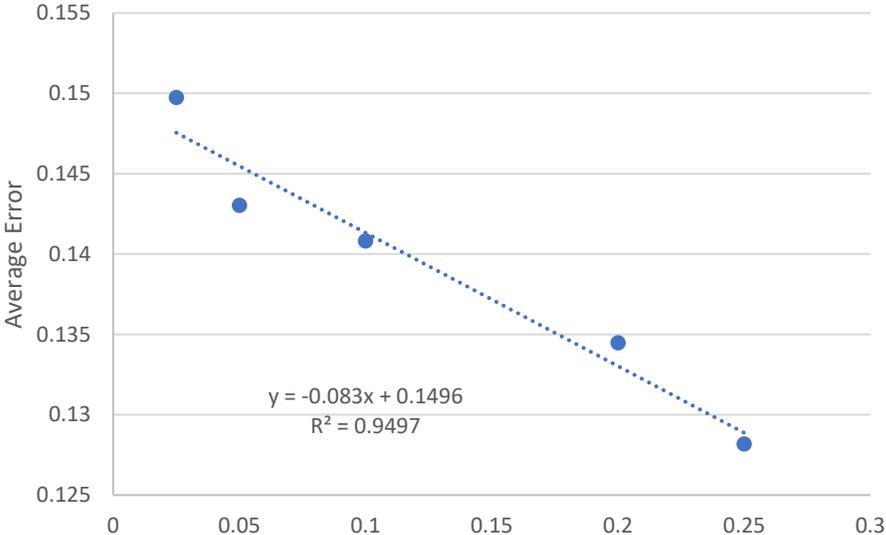

**Figure 15.** Average valid error by training velocity.

Given the demonstrated utility of increasing both the training velocity and the number of training iterations, the two changes were combined to assess the impact of concurrently altering both.

The data in Table 9 shows that, within each number of training iterations, increasing velocity reduces error. However, this doesn't resolve the issue with stopping at an arbitrary number of iterations. Thus, the pattern of increasing error between the 100 and 100, and 100 and 250 iteration levels is still present.

**Table 9.** Comparison of Training Iterations vs. Training Velocity on Average Error

| Training Iterations | Training Velocity | Average Error |
|---|---|---|
| 10 | 0.025 | 0.135760396 |
| 10 | 0.05 | 0.131683917 |
| 10 | 0.1 | 0.128124334 |
| 10 | 0.2 | 0.129051653 |
| 10 | 0.25 | 0.133989369 |
| 100 | 0.025 | 0.149748806 |
| 100 | 0.05 | 0.143028564 |
| 100 | 0.1 | 0.140812774 |
| 100 | 0.2 | 0.134484665 |
| 100 | 0.25 | 0.128187673 |
| 250 | 0.025 | 0.157499082 |
| 250 | 0.05 | 0.142105481 |
| 250 | 0.1 | 0.142707331 |
| 250 | 0.2 | 0.121074239 |
| 250 | 0.25 | 0.122565837 |
| 500 | 0.025 | 0.143163196 |
| 500 | 0.05 | 0.135429285 |
| 500 | 0.1 | 0.127506103 |
| 500 | 0.2 | 0.136924241 |
| 500 | 0.25 | 0.13354609 |

*5.3. Trigger Thresholds*

Trigger thresholds were introduced to provide a capability similar to the activation functions used in neural networks. This section characterizes their impact. Trigger thresholds, which supply a lower and upper bound for applying the result of a rule, were configured in the simulations in four different ways.

The first setting is that any result from the rule calculation is accepted through setting the thresholds at 0 and 1. This means that the thresholds are, effectively, not applied and any value from 0 to 1 will cause any actions that are attached to the rule to be fired, and the output fact will be set to the result of that rule's calculation.

The second setting was to set the thresholds to 0.2 to 0.8. Thus, any result between those two numbers would be accepted, trigger actions and be set as the value of the applicable output fact.

The third setting instructed the system to generate a random set of (valid, meaning that the upper threshold was higher than the lower one) thresholds. There could be any distance between the two numbers, as long as the upper number is not less than the lower number.

The last setting also called for a random set of thresholds to be generated; however, it ensured that there was at least a difference of 0.5 between the two numbers.  This would allow a large range of results to pass the thresholds.

The same testing protocol, as has been used in previous sections, was used for this analysis.  The results from this experimentation are shown in Table 10 and Figure 16.

**Table 10.** Average error by trigger index value.

| Trigger Index | Threshold Range | Average Error |
|---|---|---|
| 0 | 0-1 | 0.153814301 |
| 1 | .2-.8 | 0.125143707 |
| 2 | Random | 0.203840182 |
| 3 | Random with at least 0.5 difference | 0.128124334 |

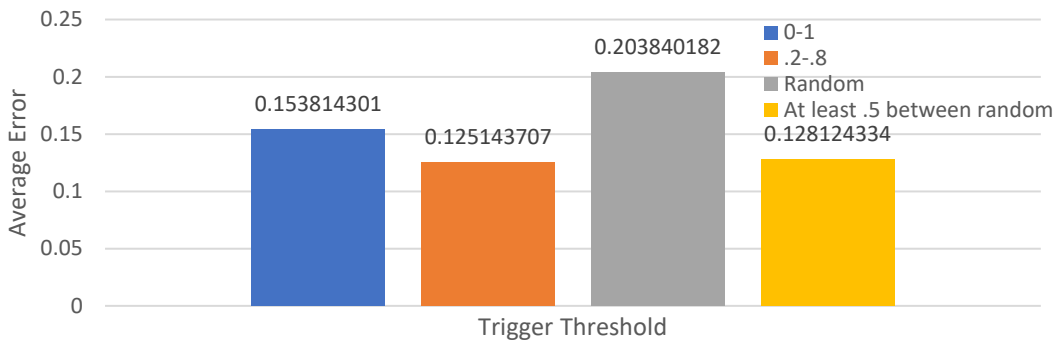

**Figure 16.** Average valid error by trigger threshold.

Notably, while the threshold values may have application-specific implications, this analysis seeks to characterize their impact generally.  The analysis of additional application-specific implications remains a topic for potential future work.  From this experimentation, it was shown that the random threshold range had the largest error among the different experiments. The lowest error was produced by the simulations with the 0.2 to 0.8 range.  The random thresholds with at least 0.5 separation produced the second-best error levels (and this error was only slightly higher than the 0.2 to 0.8 range error level).

The general impact of the thresholds was notable.  Specifically, the data shows that removing data at the extremes reduces error, even without an application-specific need for doing so. It, thus, represents a general technique for system enhancement.  This seems counter-intuitive, as it could be expected that allowing all change values to be applied would result in the lowest error, as this would allow the initial value for the start fact to have the largest impact on nodes in the network. This result demonstrates a benefit paralleling the benefit that activation functions providein neural networks.  Notably, while some restriction on when values are applied is shown to be beneficial, having an extremely small range that is applied results in less rule outputs being applied. This explains the inferior performance of the non-guaranteed-separation random approach.

*5.4. Runtime and Error Correlation Analysis*

A surrogate metric for accuracy could be beneficial for applications where error cannot be directly assessed. To this end, assessment was performed to see if the runtime of a network could be indicative of its effectiveness. If higher or lower-performing networks took different amounts of time to operate, this could be a useful method for predicting system accuracy.

However, it was found that the time it takes to run a network has a negligible correlation with the error level of that network. Table 11 and Figure 17 show data for the initial run's runtime and error levels and Table 12 shows data for the runtimes of runs that are completed after network training.

Since networks complete the same processes, irrespective of other factors, this data shows that additional time from processes or processing iterations does not necessarily produce superior or inferior error levels. Additionally, as large and complex networks use the same underlying mechanisms during a run as simple networks, the errors levels between different sizes of networks are also similar. There is a notable difference between the error values for the highest runtime levels. The error levels at these runtimes are much lower than at other levels. This is believed to indicate some complex networks with comparatively lower error; however, given that this is an extrema of the data, its analytical value is limited.

**Table 11.** Average error by initial runtime.

| Initial Runtime | Average Error |
|---|---|
| 0-1000 | 0.160983558 |
| 3000-4000 | 0.143806034 |
| 4000-5000 | 0.157649338 |
| 5000-6000 | 0.174529184 |
| 6000-7000 | 0.185296416 |
| 7000-8000 | 0.220871161 |
| 8000-9000 | 0.200991707 |
| 11000-12000 | 0.092020588 |
| 42000-43000 | 0.078693022 |
| 195000-196000 | 0.0750262 |

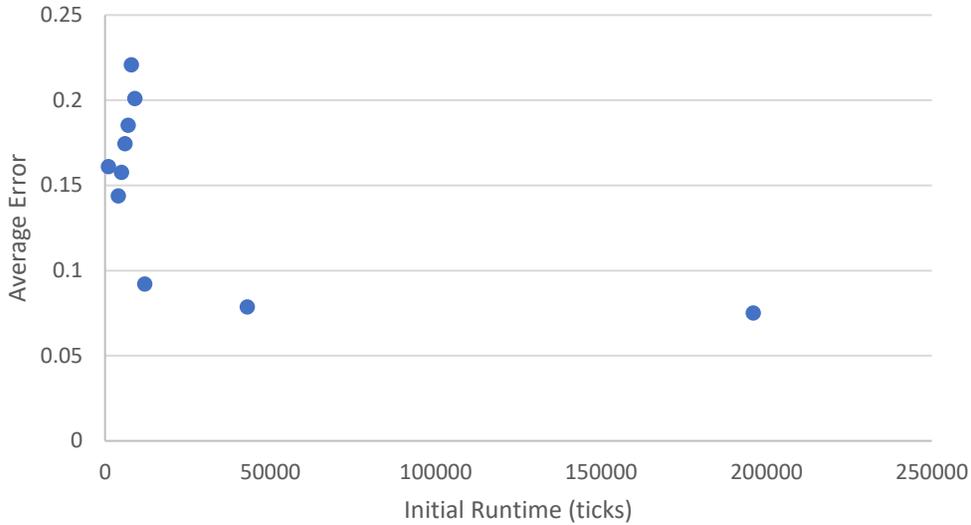

**Figure 17.** Average error by initial runtime.

**Table 12.** Average error by training time.

| Train Time | Average Error |
|---|---|
| 0-1000 | 0.160983558 |
| 3000-4000 | 0.143806034 |
| 4000-5000 | 0.156408948 |
| 5000-6000 | 0.17368452 |
| 6000-7000 | 0.1887294 |
| 7000-8000 | 0.222238306 |
| 8000-9000 | 0.177010817 |
| 11000-12000 | 0.092020588 |
| 43000-44000 | 0.078693022 |
| 194000-195000 | 0.0750262 |

Another potential way to predict system performance is using the amount of simulation runtime spent on training. Thus, focus now turns to seeing if this has a correlation with performance. The train percent metric, which is the average time of a training session divided by the same network's initial run time, is used to assess this.

The data shown in Table 13 doesn't show a clear correlation; however, there is a general negative correlation trend shown at the higher value levels that was also present in other data. Notably, the first train percent values show an upward trend, with the highest average error occurring at the third train percent level and the lowest average error occurring at the lowest train percent setting. Thus, while it seems intuitive that a longer training time will result in more rules being adjusted leading to a more accurate output value, the data suggests that the percentage of time spent on training is not a useful predictor of system performance.

**Table 13.** Train Percent vs. Average Error

| Train Percent | Average Error |
|---|---|
| 0.45-0.5 | 0.137549715 |
| 0.8-0.85 | 0.159895705 |
| 0.85-0.9 | 0.185505253 |
| 0.95-1 | 0.142501633 |
| 1-1.05 | 0.167174342 |
| 1.05-1.1 | 0.150878543 |
| 1.1-1.15 | 0.14319422 |

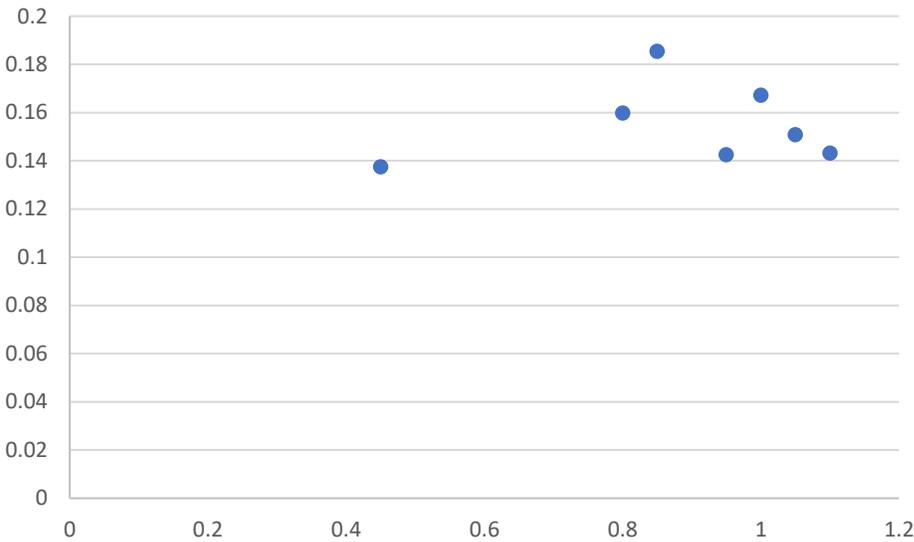

**Figure 18.** Average error by training percent.

*5.5. Impact of Combined Settings Changes*

This section considers the impact of making several settings changes concurrently. First, the impact of varying the number of training iterations and actions concurrently is assessed. Next, the impact of varying the number of training iterations and the action fire likelihood is considered. Finally, assessment turns to varying the number of training iterations and the trigger threshold concurrently.

Table 14 shows that, within each level of training iterations, there is only a consistent increase of the average error with an increased number of actions, with a single outlier / anomaly. Generally, thus, error increases with both additional training iterations and actions and increases at a higher rate when both are concurrently increased.

As was previously discussed, error when stopping at an arbitrary iteration increases with the number of iterations; however, the best performance of the network typically improves with increased training. Given this, best performance and action impact would appear to exhibit opposite trends.

**Table 14.** Average error for different levels of training iterations and action count.

| Training Iterations | Action Count | Average Error |
|---|---|---|
| 10 | 50 | 0.141485 |
| 10 | 100 | 0.166171 |
| 10 | 250 | 0.182415 |
| 10 | 500 | 0.177011 |
| 100 | 50 | 0.156095 |
| 100 | 100 | 0.18303 |
| 100 | 250 | 0.201467 |
| 100 | 500 | 0.215266 |
| 250 | 50 | 0.158295 |
| 250 | 100 | 0.189636 |
| 250 | 250 | 0.200507 |
| 250 | 500 | 0.224973 |
| 500 | 50 | 0.170366 |
| 500 | 100 | 0.19152 |
| 500 | 250 | 0.214281 |
| 500 | 500 | 0.226477 |

Table 15 shows the error level with different levels of training and different action fire probabilities. When a rule is run, each action associated with that rule has a probability of being fired. While normally all actions of a triggered rule would run, this probability represents the likelihood of a rule's run providing data that changes a fact's value, which will vary by application area. When an action is fired, a random fact's value is changed to a random value.

The data shows that average error increases with both training level and action fire probability. This behavior is consistent with increasing error with greater numbers of actions and higher training, as just discussed.

**Table 15.** Average error for different levels of training iterations and action fire probability.

| Training Iterations | Action Fire Probability | Average Error |
|---:|---:|---:|
| 10 | 0.25 | 0.132574 |
| 10 | 0.5 | 0.144514 |
| 10 | 0.75 | 0.145573 |
| 10 | 1 | 0.166171 |
| 100 | 0.25 | 0.140534 |
| 100 | 0.5 | 0.162374 |
| 100 | 0.75 | 0.168755 |
| 100 | 1 | 0.18303 |
| 250 | 0.25 | 0.13489 |
| 250 | 0.5 | 0.159345 |
| 250 | 0.75 | 0.182457 |
| 250 | 1 | 0.189636 |
| 500 | 0.25 | 0.142403 |
| 500 | 0.5 | 0.164446 |
| 500 | 0.75 | 0.161463 |

| | 500 | 1 | 0.19152 |
|---|---|---|---|

Table 16 shows the error produced by different training levels and different thresholds settings. As already discussed, the number of training iterations has a positive correlation with the average error. Within each training level, a pattern between the different trigger thresholds holds consistently. The random threshold always performs the worst and the 0.2-0.8 always performs the best, with the other two falling in between.

**Table 16.** Average error for different levels of training iterations and trigger threshold settings.

| Training Iterations | Trigger Threshold Range | Average Error |
|---|---|---|
| 10 | 0-1 | 0.167535 |
| 10 | .2-.8 | 0.160182 |
| 10 | Random | 0.17833 |
| 10 | Random with at least 0.5 difference | 0.166171 |
| 100 | 0-1 | 0.192606 |
| 100 | .2-.8 | 0.182436 |
| 100 | Random | 0.203177 |
| 100 | Random with at least 0.5 difference | 0.18303 |
| 250 | 0-1 | 0.185096 |
| 250 | .2-.8 | 0.170537 |
| 250 | Random | 0.212537 |
| 250 | Random with at least 0.5 difference | 0.189636 |
| 500 | 0-1 | 0.196888 |
| 500 | .2-.8 | 0.188036 |
| 500 | Random | 0.205806 |
| 500 | Random with at least 0.5 difference | 0.19152 |

Overall, the combination of different conditions has demonstrated basically the results that would be expected based on the conditions' individual performance. This is demonstrative of the stability of the performance of the system.

**6. Conclusions and Future Work**

This paper has proposed a system which provides a rule-fact based autonomous control capability, which is based on the combination of the Blackboard Architecture and the defensible gradient descent-trained expert system. This system is designed to be able to be trained and operated in a manner similar to its predecessors while providing several capabilities that exceed those of the systems presented in prior work.

Specifically, the Blackboard Architecture action object was added to the gradient descent-trained expert system to provide an actualization capability. An alternate (to prior work) path-based optimization algorithm was implemented and demonstrated to perform sufficiently well for many applications. Also, a threshold-based activation function was implemented and demonstrated to improve general performance, in addition to providing potential support for specific application areas which require this functionality.

Like the gradient descent trained expert system, the system proposed herein provides precise control over what decision pathways can be established, as the training process cannot create new ones and only optimizes the ones provided. This capability is perhaps even more important for this system, as compared to the expert system, given its actualization capabilities.

Simulations have been run to demonstrate the efficacy of this system and characterize its performance under a variety of experimental conditions. A number of key conclusions can be drawn from this experimentation. First, the results of the simulations demonstrate that the system's core capabilities function and operate consistently with its progenitor systems' operations. Second, the use of random networks with different characteristics has demonstrated the general potential of this system. The data generated from this analysis also provides key inputs to aid system designers seeking to implement the proposed functionality for application areas.

This paper also introduced the use of upper and lower threshold values to provide an activation function. While this was demonstrated in the context of the gradient descent-trained Blackboard Architecture system presented herein, it can also be applied to gradient descent-trained expert systems. These thresholds were shown to influence network operations, suggesting their utility for a variety of potential application-specific uses. The 0.2 – 0.8 threshold was shown to have a general performance-enhancing capability and may be beneficial to use, even in the absence of an application-driven need.

The experimentation also demonstrated that, as is typical with neural networks and was previously demonstrated with the gradient-descent expert system, training has a notable impact on the results of a network run. Both the number of training iterations and the velocity at which the system is trained were shown to have an influence. This process demonstrated the efficacy of the newly proposed path-based training algorithm, which modifies a lower number of nodes during the training process and, thus, will provide benefits in terms of reducing the required level of memory (or disk) access. This algorithm was shown to very easily depart from optimal configurations and a mechanism that was added to retain the best result was shown to provide a significant error-reduction performance benefit.

Fundamentally, the system described herein is a defensible, human-understandable, and learning-capable Blackboard Architecture implementation. Like its predecessors, it operates using a rule-fact-action network. It uses decimal values for rules and contribution processing for facts, like the gradient descent-trained expert system, and provides the actualization capabilities of the Blackboard Architecture. Using these simple and inherently understandable capabilities, extremely complex decisions can be made. While large networks can be created to solve complex challenges, the simple calculations and structure of the system makes the reason for those decisions inherently traceable and understandable by human operations.

Future work in this area will include the evaluation of the efficacy of the use of these new capabilities in various application areas. This will aid efforts in the targeted area as well as facilitating the continued characterization and efficacy demonstration of the proposed system. The development of other optimization algorithms and activation functions is another key area of potential future work. This includes the comparative evaluation of these algorithms for various application areas to assess if performance is closely coupled with network design (and, thus, application implementation) or not.